\documentclass[a4paper,conference]{IEEEtran}

\usepackage{subfig}
\usepackage{times}
\usepackage{epsfig}
\usepackage{amsmath}
\usepackage{amssymb}

\usepackage{csquotes}
\usepackage{multirow}
\usepackage{colortbl,xcolor}
\usepackage{booktabs}
\usepackage{mathtools}
\usepackage{textpos}
\usepackage{breqn}

\definecolor{customgray}{rgb}{0.9, 0.9, 0.9}
\newcolumntype{g}{>{\columncolor{customgray}}c}
\newcolumntype{z}{>{\columncolor{customgray}}l}
\newcolumntype{?}[1]{!{\vrule width #1}}

\begin{document}
%
\title{Efficient Action Recognition \\ Using Confidence Distillation}

\author{
\IEEEauthorblockN{Shervin Manzuri Shalmani}
\IEEEauthorblockA{Department of Computing \\ and Software\\
McMaster University\\
Ontario, Canada\\
Email: manzuris@mcmaster.ca}
\and
\IEEEauthorblockN{Fei Chiang}
\IEEEauthorblockA{Department of Computing \\ and Software\\
McMaster University\\
Ontario, Canada\\
Email: fchiang@mcmaster.ca}
\and
\IEEEauthorblockN{Rong Zheng}
\IEEEauthorblockA{Department of Computing \\ and Software\\
McMaster University\\
Ontario, Canada\\
Email: rzheng@mcmaster.ca}
}


%


\maketitle

\begin{abstract}
  Modern neural networks are powerful predictive models. However, when it comes to recognizing that they may be wrong about their predictions, they perform poorly. For example, for one of the most common activation functions, the ReLU and its variants, even a well-calibrated model can produce incorrect but high confidence predictions. Most current action recognition methods are based on clip-level classifiers that densely sample a given video for non-overlapping, same-sized clips and aggregate the results using an aggregation function - typically averaging - to achieve video level predictions. While this approach has shown to be effective, it is sub-optimal in recognition accuracy and has a high computational overhead. To mitigate both these issues, we propose the \textit{confidence distillation framework} to teach a student model how to select less ambiguous clips for the teacher, and divide the task of prediction between the two. We conduct extensive experiments on three action recognition datasets and demonstrate that our framework achieves significant improvements in action recognition accuracy (up to $20\%$) and computational efficiency (more than $40\%$).
\end{abstract}

%
\IEEEpeerreviewmaketitle

\section{Introduction}
With the ever-increasing amount of videos being captured, shared and consumed every day, it is imperative to design systems that can efficiently analyze this content without a high computational cost. Through recent advances in action recognition, robust classifiers have been designed \cite{Hara_2018_CVPR, tran2015learning, tran2018closer, tran2019video, feichtenhofer2017spatiotemporal, feichtenhofer2019slowfast, feichtenhofer2020x3d} that operate on short time spans of the video - typically spanning a few seconds - as input clips and generate video-level classification by applying an aggregation operator - typically averaging - to all clip-level predictions over the entire video; Applying a clip classifier over all individual clips of a video stream may be reasonable when the video length is known to be short. Unfortunately, such an assumption does not hold in real-world applications and becomes impractical. Videos may be more than a few seconds long, providing the classifier with ample opportunity to misclassify individual clips, negatively affecting the aggregate video-level classification accuracy. To mitigate this issue, clips can be sampled randomly, at equidistant intervals or take the redundancy of the video into account and skip redundant frames. Our fundamental intuition is that no matter which of the previous approaches is taken, the aleatoric uncertainty of data will still provide aggregation-based models with opportunities to misclassify. 

\begin{figure}[t] %
    \centering
    {{\includegraphics[width=1\linewidth]{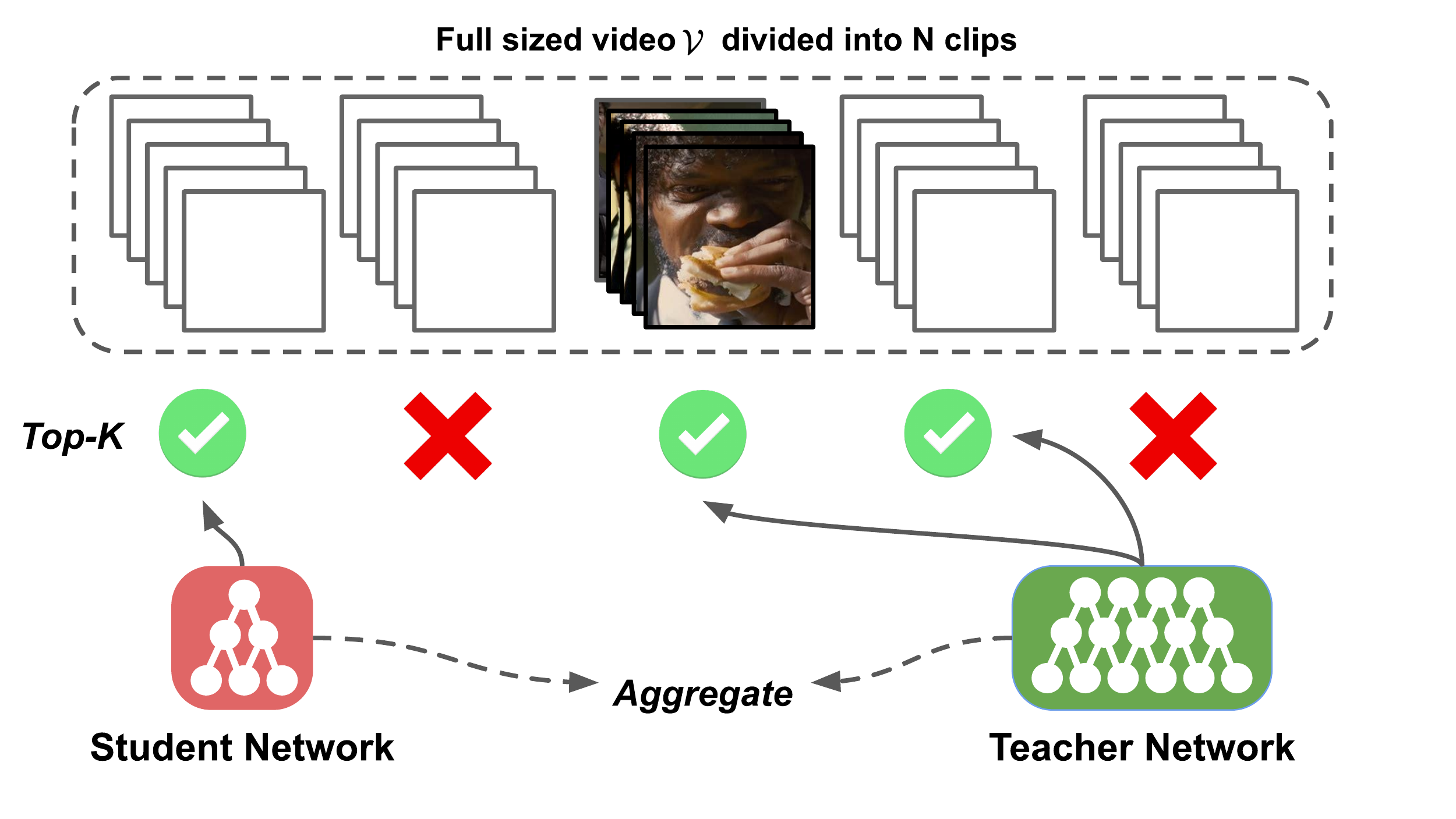}}}%
    \caption{An illustration of the confidence distillation inference framework. The student sampler learns to skim the videos and select the $K$ least ambiguous clips, that are then divided for prediction between the student and the teacher.}%
    \label{fig:abstframework}
\end{figure}

We address both challenges of computational efficiency and aggregation accuracy by taking into account the \textit{ambiguity} of individual clips. Many current state-of-the-art approaches to clip sampling assume that the highest probability score in the network's output indicates high confidence and low ambiguity \cite{korbar2019scsampler, gao2020listen, wu2019adaframe}. 

However, when deployed in the real world, machine learning models \cite{goodfellow2014explaining, amodei2016concrete} and ReLU networks specifically \cite{hein2019relu} often fail silently by providing high confidence predictions (giving a high probability to one class) while being incorrect. Looking at Fig. \ref{fig:abstframework}, this means some subsequent clips in the same video that looks similar to the human eye can make the classifier fail due to inherent noise. Motivated by this fact, in this work, we propose the "Confidence distillation training and inference framework," a novel distillation scheme where we propose a loss function to train a lightweight sampler with the primary function of skimming the video for the least ambiguous clips and the secondary function of classifying "easier" clips at low complexity. In other words, the student learns to output both a classification score and a score $\tilde{z}$ for its confidence in the teacher's ability to correctly classify that specific input. We will only perform prediction on the best $K$ clips using either only the teacher model - which reduces the computational cost and increases accuracy - or divide the prediction task between the student and the teacher, to achieve higher computation savings at little cost to accuracy. Our key contributions are summarized as follows.

\begin{itemize}
    \item We propose a novel distillation loss (Eq. \eqref{lossfunc}) to train a lightweight classifier that also learns to predict which clips will likely be ambiguous for the teacher (Fig. \ref{fig:inference}). 
    \item We conduct extensive experiments to show that this framework outperforms state-of-the-art models in detecting ambiguous clips, significantly increases accuracy (up to 20\% increase in accuracy) and notably increases computational efficiency (more than 40\% faster inference) across three real-world video datasets of varying lengths, the UCF-101, Kinetics and Something-something V2.
\end{itemize}

\section{Related work}

{\bf Action recognition:} Recent architectures are mostly designed by extending image classifiers with a temporal dimension while preserving the spatial properties of each frame. Among them include directly transforming 2D models \cite{simonyan2014very, krizhevsky2012imagenet, he2016deep, szegedy2015going} such as Inception or ResNet to 3D \cite{carreira2017quo, Hara_2018_CVPR}, adding RNNs on top of 2D CNNs \cite{donahue2015long, li2018recurrent, li2018videolstm, yue2015beyond, sun2017lattice, yosinski2015understanding}, using two identical 2D CNNs or using more sophisticated volume-based convolutions \cite{feichtenhofer2017spatiotemporal,qiu2017learning, taylor2010convolutional, tran2015learning, tran2019video, tran2018closer}. The work, as mentioned earlier, focuses on building powerful models to improve action recognition without considering the computational cost, whereas we aim to perform action recognition accurately and efficiently.

{\bf Efficient architectures:} Many innovative architectures have been proposed for efficient video classification \cite{kopuklu2019resource, zhu2020faster, zolfaghari2018eco, wu2018compressed, wang2018appearance, wang2020video, tran2019video, sun2018optical, piergiovanni2019representation, luo2019grouped, lee2018motion, hussein2019timeception, fan2018end, diba2018spatio, feichtenhofer2020x3d}. The idea behind them is to optimize the classifier architecture itself, while in contrast, our framework is mostly architecture-independent. In \cite{feichtenhofer2020x3d, lin2019tsm}, a given 2D architecture is extended across as temporal dimension. Efficient architectures \cite{howard2017mobilenets, ma2018shufflenet} using depth-wise and channel-wise separable convolutions have recently been applied \cite{kopuklu2019resource, tran2019video} to video processing. These approaches are orthogonal to ours and can be applied alongside the proposed method.

{\bf Clip and frame selection:} These methods aim to train a model to do the video clip or frame sampling \cite{alwassel2018action, korbar2019scsampler, su2016leaving, wu2019multi, wu2019adaframe, yeung2016end, gao2020listen}. SCSampler \cite{korbar2019scsampler} uses compressed features with \textit{softmax confidence} scores to score clips for their visual sampler and selects the top $k$ clips. Similarly, in IMGAUD2VID \cite{gao2020listen} multiple modalities such as audio and video are used to select less redundant frames. Other state-of-the-art include using reinforcement learning \cite{wu2019adaframe} training single and \cite{wu2019multi} multiple agents to perform frame selection collaboratively. By contrast, our method requires neither a complex RL policy gradient nor access to audio-level features.
Furthermore, in all methods, the video is assumed to be very long and redundant, and the number of sampled clips can be large. By focusing on redundancy, the effectiveness of these models diminishes as the size of the dataset becomes shorter and there are fewer irrelevant sections in the video. Nevertheless, these methods are complementary to ours. In our method, we do not make an assumption on video length, and as such, we can complement our sampler by first applying \cite{gao2020listen, korbar2019scsampler, wu2019adaframe} to a long video to reduce the search space and then sample a subset of those frames with confidence distillation. Finally, most methods use the \textit{softmax confidence} score as a proxy for clip ambiguity, which is not a good surrogate for classifier confidence. Therefore, we make no such assumption and assume that the classifier can be confidently incorrect. 

{\bf Confidence estimation:} The task of determining when a network should say \textit{\enquote{I do not know.}} or provide confidence estimates for its outputs. Examples include Bayesian neural networks \cite{mackay1992bayesian}, variational inference \cite{blundell2015weight, farquhar2020radial} and simpler methods such as Monte Carlo dropout \cite{gal2016uncertainty} where measures such as \textit{predictive entropy} are used. Other methods learn unsupervised, unbounded \cite{kendall2017uncertainties} or bounded confidence values \cite{devries2018learning} for classification and regression. In \cite{gal2016uncertainty, hein2019relu} \textit{Predictive entropy} is a measure applied to capture aleatoric uncertainty; the type of uncertainty inherent in the data such as an image of a number 7 that may appear similar to a number 1. In \cite{devries2018learning}, a given untrained classifier is regularized during training to detect out-of-distribution samples without supervision. We take a different approach by having the student sampler learn confidence estimates for its teacher's output to improve computational efficiency and accuracy in the video analysis task.

{\bf Other related works:}
Compression techniques such as knowledge distillation \cite{bucilua2006model, hinton2015distilling, urban2016deep} address the problem of distilling the knowledge of large pre-trained models into smaller, more efficient models, which tends to perform better than having the smaller model learn from raw data \cite{furlanello2018born, cheng2020explaining}. The student is typically deployed without the teacher. Using our framework, we can optionally employ both the student sampler and the teacher to address the problem of efficient video analysis cooperatively. 

\section{Approach}
\label{approach}

Our goal is to perform more accurate and efficient action recognition in videos by skipping over ambiguous clips. We will formally define our problem (Sec. \ref{problemformulation}) and introduce the motivation behind teaching the student the notion of confidence - how confident it is that the teacher will not make a mistake - using the teacher's outputs (Sec. \ref{motivation}); finally, we present how this confidence score will be used to skip over ambiguous clips in the video (Sec. \ref{algorithmdetails}).

\subsection{Problem formulation}
\label{problemformulation}

Given a video $\mathcal{V}$ of arbitrary length, the goal of video classification is to map $\mathcal{V} \in \mathbb{R}^{T \times 3 \times H \times W}$ into a fixed set of $C$ classes. Since $\mathcal{V}$ can be arbitrarily long, it is often difficult and sometimes impossible to stack all the video frames together - which can be hundreds and up to thousands - as input to a single deep network.

Due to this constraint, a majority of current approaches \cite{simonyan2014two, tran2015learning, carreira2017quo, tran2018closer, wang2018non, feichtenhofer2019slowfast} first train a clip classifier $\mathrm{F}(\cdot)$ that operates on a short fixed-length video clip of $K$ frames with size $H \times W$. Given a full-sized video, the clip-classifier is applied to all $N$ clips $\{\mathbf{V}_1, \mathbf{V}_2, \ldots, \mathbf{V}_N\}$ where $N = \frac{T}{K}$ and padding is applied to the last clip when needed. The final video-level predictions are obtained by aggregating the clip-level predictions of all $N$ clips, where the aggregation is usually average pooling.

The dense sampling paradigm is both inaccurate and inefficient. As the length of $\mathcal{V}$ grows, so does the computational cost of inferencing. It also results in poor prediction accuracy since every incorrect or ambiguous clip prediction across $\{\mathbf{V}_1, \mathbf{V}_2, \ldots, \mathbf{V}_N\}$  will negatively affect the aggregate prediction of $\mathcal{V}$. Given a pre-trained clip classifier $\mathrm{F}(\cdot)$, our goal is to train an efficient student sampler-classifier $\mathrm{f}(\cdot)$ that can predict whether $\mathrm{F}(\cdot)$, the teacher, will misclassify a given clip $\mathbf{V}_i$ and also make predictions on the easier clips. The student identifies the least ambiguous clips that can be passed to the teacher while (2) classifying the easiest clips on its own to increase the speed of inferencing further, thereby addressing both challenges at once. 

We note that during this procedure, the classifier $\mathrm{F}(\cdot)$ is left unmodified. This property renders our approach helpful as a post-training procedure to improve existing classifiers' accuracy and efficiency.

\subsection{Shared representation learning objective}
\label{motivation}

Naively, one will independently train a lightweight network $\mathrm{f}(\cdot)$ separately from the classifier $\mathrm{F}(\cdot)$ on the same training set $\mathcal{D}$ with one-hot labels. This implies that there is no guarantee that the sampler and the classifier would agree on clip ambiguity without supervision signals. In order to improve accuracy, only the clips $\mathrm{F}(\cdot)$ finds ambiguous are relevant. 

To provide supervision signals that denote ambiguity, we can design pseudo-ground-truth binary confidence labels on the training set $\mathcal{D}$ denoted  $z_i$ as follows: Given that the ground-truth classification label for each clip $V_i$ is $y$, we apply the softmax function to the prediction logits from $\Omega(V_i)$ to obtain the vector of class prediction probabilities $\mathbf{p}$. We then find which class $c$ has the maximum prediction probability $p_c$ in $\mathbf{p}$. The confidence scores $z_i$ for each clip $V_i$ are then defined as:

\begin{eqnarray}
z_i = \begin{cases} 
\begin{array}{cl}
1 & \text{ if~~ } argmax_{c \in \{1,\dots,C\}}(\mathbf{p}) = y\\
0 & \text{ otherwise}
\end{array}
\end{cases}
\label{psuedolabels}
\end{eqnarray}

In other words, a clip has a confidence score of 0 when $\mathrm{F}(\cdot)$ misclassifies it. Naively, the learning objective for $\mathrm{f}(\cdot)$ can correspond to a binary classification task on these confidence labels using a binary cross-entropy as the loss function (the task can also be formulated as a regression task, similar to \cite{korbar2019scsampler}. However, we found little difference in learning a regressor vs. learning a classifier in our experiments). Imbalanced data and lack of fine-grained supervision results in this non-trivial approach having the same performance as random sampling as seen in Table \ref{tab:classification_accuracy}.

Intuitively, when learning where one is likely to make mistakes, it also helps to learn about the context; here, the context is the video class where some clips belonging to more challenging classes can be more ambiguous than others. This prior could lead to sparser and more informative predictions \cite{paredes2012exploiting}. Using just the one-hot training labels, we only know that a given activity clip belongs to one class, such as \enquote{running} or \enquote{jumping}. Through the use of soft labels  \cite{hinton2015distilling} - i.e. other than the highest probability class, other classes also have non-zero probabilities in the categorical distribution - we can access extra information and see the probability of the activity clip belonging to each class via a ranked list of probabilities. Our intuition is that when learning to predict instances where the teacher is likely to make mistakes, this knowledge about class predictions may be helpful in a multi-task learning \cite{caruana1997multitask} objective where a shared representation is learned for multiple objectives.

In contrast to the naive learning objective, learning a shared representation using knowledge distillation utilizes the softened output of a trained \textit{teacher} network that contains more information about a data point than the one-hot class label. e.g., if multiple classes are assigned high probabilities for a video clip, it may lie close to a decision boundary among those classes. 

Concretely, given an input video clip $\mathbf{V}$, the teacher network $\mathrm{F}(\cdot)$ produces a vector of logits $\mathbf{s}^t(\mathbf{V})$:

\begin{eqnarray}
\mathbf{s}^t(\mathbf{V}) = [s^t_1(\mathbf{V}), s^t_2(\mathbf{V}), \ldots, s^t_C(\mathbf{V})]
\end{eqnarray}

In order to produce \enquote{softened}, non-peaky and more informative probability distributions from $\mathbf{s}^t(\mathbf{V})$, temperature scaling is used alongside the softmax function \cite{hinton2015distilling, guo2017calibration} to produce $\tilde{\mathbf{p}}^t(\mathbf{V})$:

\begin{align}
\tilde{\mathbf{p}}^t(\mathbf{V}) &= [\tilde{p}^t_1(\mathbf{V}), \tilde{p}^t_2(\mathbf{V}), \ldots, \tilde{p}^t_C(\mathbf{V})]\\
    \text{where}\quad \tilde{p}^t_k(\mathbf{V}) &= \frac{e^{s^t_k(\mathbf{V})/\tau}}{\sum_j e^{s^t_j(\mathbf{V})/\tau}}
\end{align}

where $\tau$ is the temperature hyperparameter. The student model $\mathrm{f}(\cdot)$ similarly produces a softened class probability distribution, $\tilde{\mathbf{p}}^s(\mathbf{V})$. The student also needs to learn a confidence score $\tilde{z}^s(\mathbf{V}) \in [0,1]$ using the pseudo-ground-truth binary labels from Eq. \ref{psuedolabels}. This presents an interesting optimization issue: When the student is learning to classify, we want the overall loss to decrease when it has correctly classified the input; but in cases where the confidence score of a video clip is 0, $\mathbf{p}^t(\mathbf{V})$ as provided by the teacher would be misleading. To mitigate this issue, we modify the loss function similar to \cite{kendall2017uncertainties, gurevich2018pairing, devries2018learning}, and guide the student by incorporating the confidence scores into the distillation loss. 

The student must have confidence that the teacher is right to learn from the teacher. When the student thinks that the teacher may be wrong, they can ask for more information; this is akin to increasing $\tau$ when the predicted confidence is low. When $\tau \rightarrow \infty$, the pseudo-probability output classes end up with a uniform distribution. When the $\tilde{z} = 0$, we need $\tau \rightarrow \infty$ and when $\tilde{z} = 1$, we need $\tau \rightarrow T$, where $T$ is also a hyperparameter.

For a given video clip $\mathbf{V}$ we further modify the teacher's probability scores after applying the softened softmax to $\tilde{\mathbf{p}}^t(\mathbf{V})$ as follows:

\begin{eqnarray}
\thinmuskip=\muexpr\thinmuskip*5/8\relax
\medmuskip=\muexpr\medmuskip*5/8\relax  
\hat{\mathbf{p}}^t(\mathbf{V}) =
\begin{cases} 
\begin{array}{cl}
\hspace{-.5em}\tilde{z~} \tilde{\mathbf{p}}^t(\mathbf{V}) + (1 - \tilde{z})~ \mathcal{U}([0,1]^C)  & \text{ if~~ } z=1\\
\hspace{-.5em}(1 - \tilde{z})~ \tilde{\mathbf{p}}^t(\mathbf{V}) + \tilde{z~} \mathcal{U}([0,1]^C) & \text{ o.w.}
\end{array}
\end{cases}
\label{eq5}
\end{eqnarray}

where $\mathcal{U}([0,1]^C)$ is the uniform distribution between 0 and 1 over the $C$ classes (with the same shape as $\mathbf{\tilde{p}^t}$). In the overall loss, we include a distillation loss $\mathcal{L}_{\mathit{KD}}$ to match the student and teacher outputs. Furthermore, to prevent a naive solution of $\tilde{z} = 0$ from being converged to as the training progresses, we add a binary cross-entropy loss $\mathcal{L}_{conf}$ over the ground-truth confidence labels. The overall loss is then:

\begin{eqnarray}
\mathcal{L} = 
\mathcal{L}_{\mathit{KD}}(\hat{\mathbf{p}}^t, \tilde{\mathbf{p}}^s) + \lambda \mathcal{L}_{conf}
\label{lossfunc}
\end{eqnarray}

wherein the $\mathcal{L}_{\mathit{KD}}$ term is a knowledge distillation loss with modified logits from Eq. \ref{eq5}; $\tau$ and $\lambda$ are hyperparameters. In $\mathcal{L}_{\mathit{KD}}(\hat{\mathbf{p}}^t, \tilde{\mathbf{p}}^s)$ the similarity of the two pseudo-probability posterior distribution vectors for input $\mathbf{V}$ of the student and the teacher is measured using the KL divergence metric:

\begin{eqnarray}
\mathcal{L}_{\mathit{KD}}(\hat{\mathbf{p}}^t, \tilde{\mathbf{p}}^s) = 
\tau^2 [ -\frac{1}{n} \sum_{i=1}^{n} \hat{p}^t_i(\mathbf{V}) \mathbf{log}\frac{\tilde{p}_i^s(\mathbf{V})}{\hat{p}_i^t(\mathbf{V})} ] 
\end{eqnarray}

which is equal to the difference between the cross entropy of the labels $H(\hat{\mathbf{p}}^t, \tilde{\mathbf{p}}^s)$ and the empirical entropy $H(\hat{\mathbf{p}}^t)$:

\begin{eqnarray}
\mathcal{L}_{\mathit{KD}}(\hat{\mathbf{p}}^t, \tilde{\mathbf{p}}^s) = H(\hat{\mathbf{p}}^t, \tilde{\mathbf{p}}^s) - H(\hat{\mathbf{p}}^t)
\end{eqnarray}

We refer to this loss variant \textit{ConDi-SR} (confidence distillation using a shared representation).

\subsection{Inferencing}
\label{algorithmdetails}

Once $\mathrm{f}(\cdot)$ has been trained using the aforementioned loss variants, we can utilize $\mathrm{f}(\cdot)$ to sample the top $K$ clips in $\mathbf{V}$ to be classified by $\mathrm{F}(\cdot)$.

On the other hand, in the case of \textit{ConDi-SR}, we have more options other than feeding the top $K$ clips. Since $\mathrm{f}(\cdot)$ also learns to classify in a separate branch, it can also be utilized to classify easier clips during inference and send only the more challenging clips to $\mathrm{F}(\cdot)$ for increased efficiency. This results in two ranked lists, one sorted by the student's confidence that the teacher will be right on that clip and the other sorted by the uncertainty of the student over its own prediction, using the predictive entropy of the classification output. We use a hyperparameter $K_s$ to merge these two lists, denoting the number of ranked clips to be predicted by the student sampler, while the rest $K_t = K - K_s$ is given to the teacher.  

Note that while \textit{ST-Conf} also learns a shared representation of confidence and classification, it differs from \textit{ConDi-SR}. In \textit{ST-Conf}, the confidence scores represent the \textit{confidence of the model itself} while in \textit{ConDi-SR} the confidence scores represent the \textit{confidence of the student in correct classification by the teacher}. This allows \textit{ConDi-SR} to be able to produce two meaningful ranked lists. 

\begin{figure*} %
    \centering
    \subfloat[\centering Training]
    {{\includegraphics[width=0.3\linewidth]{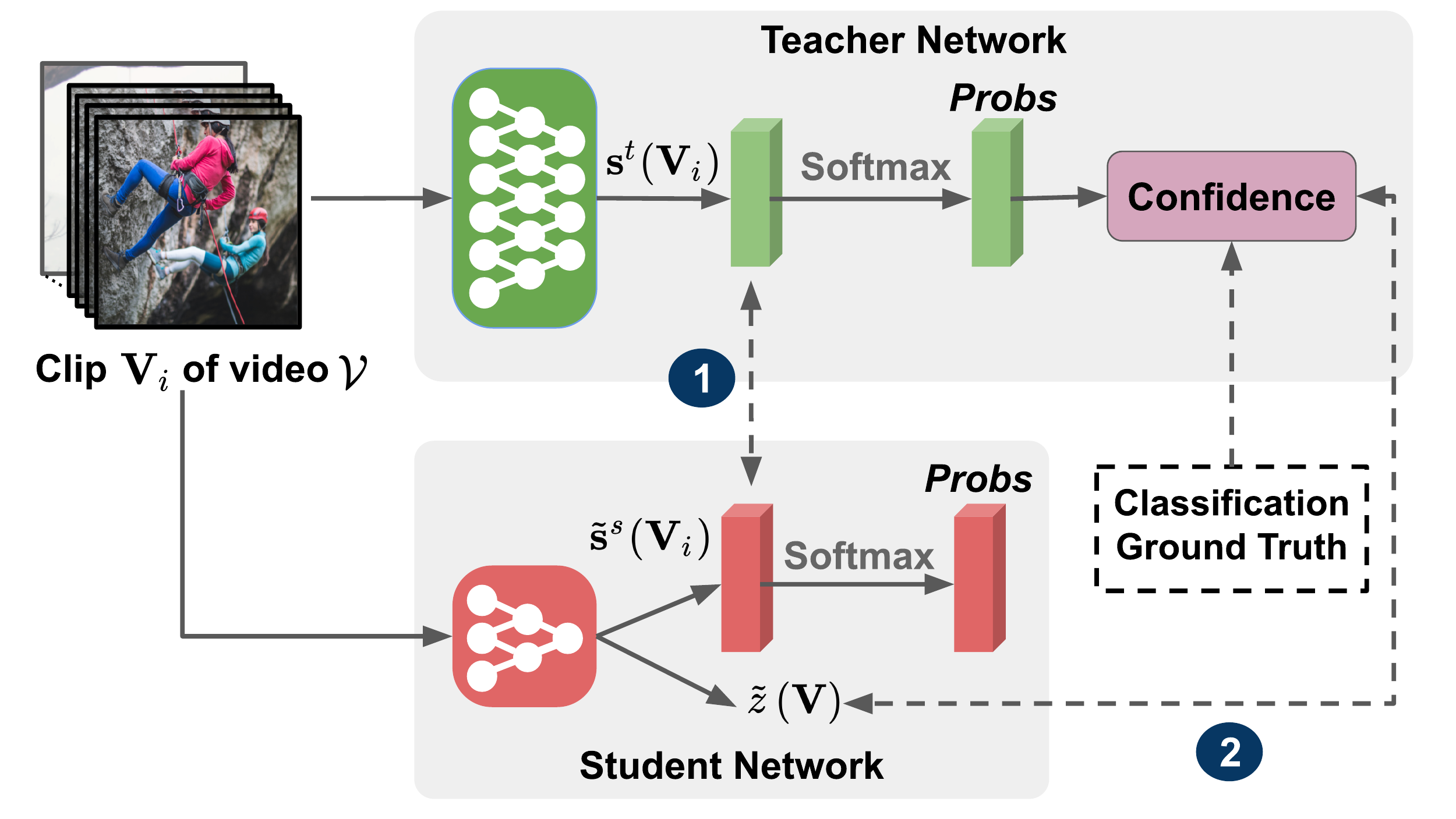}}}%
    \quad
    \vrule
    \quad
    \subfloat[\centering Inference using only the confidence scores.]
    {{\includegraphics[width=0.3\linewidth]{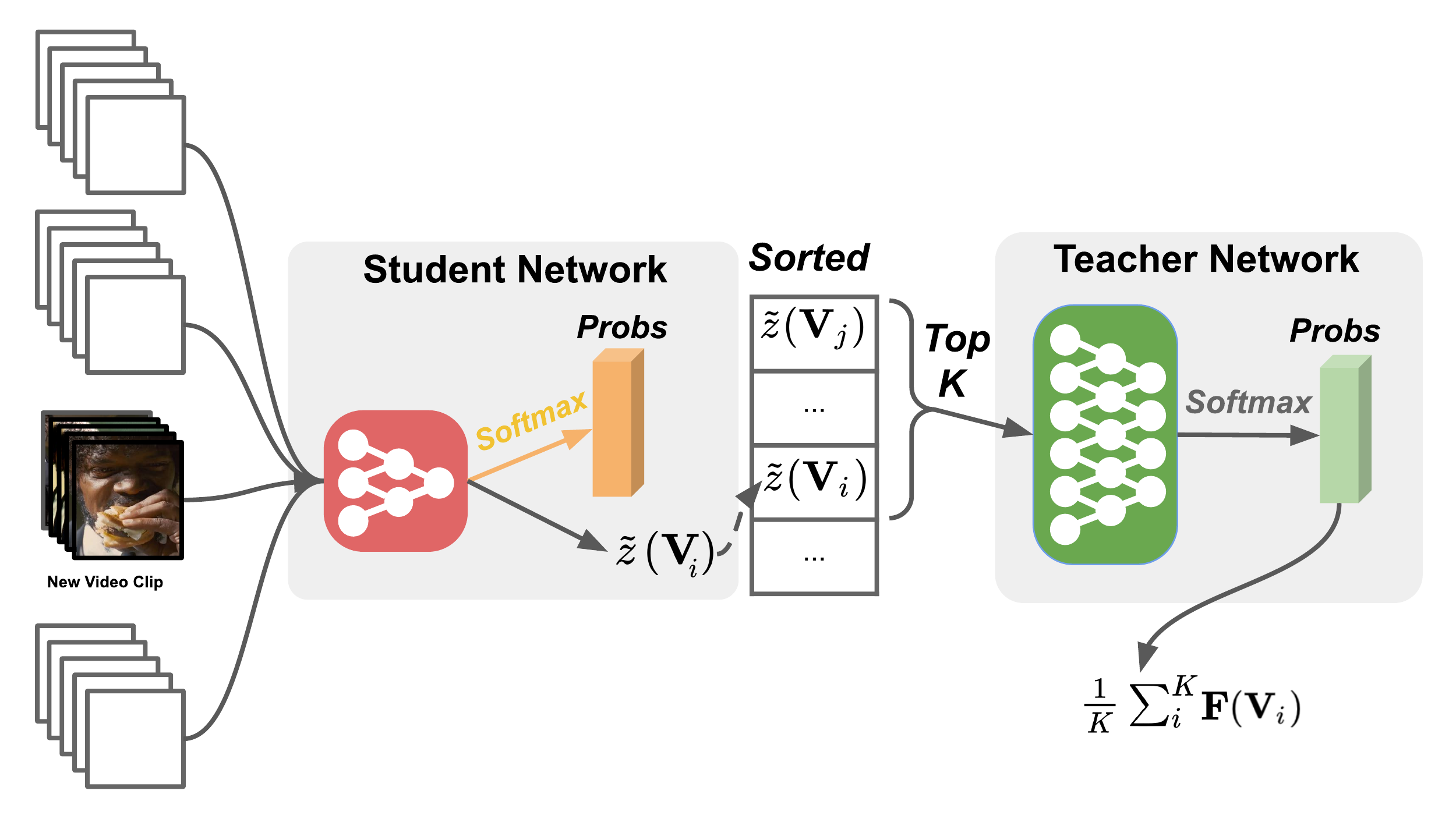}}}%
    \qquad
    \subfloat[\centering Inference using two ranked lists produced by \textit{ConDi-SR}.]
    {{\includegraphics[width=0.3\linewidth]{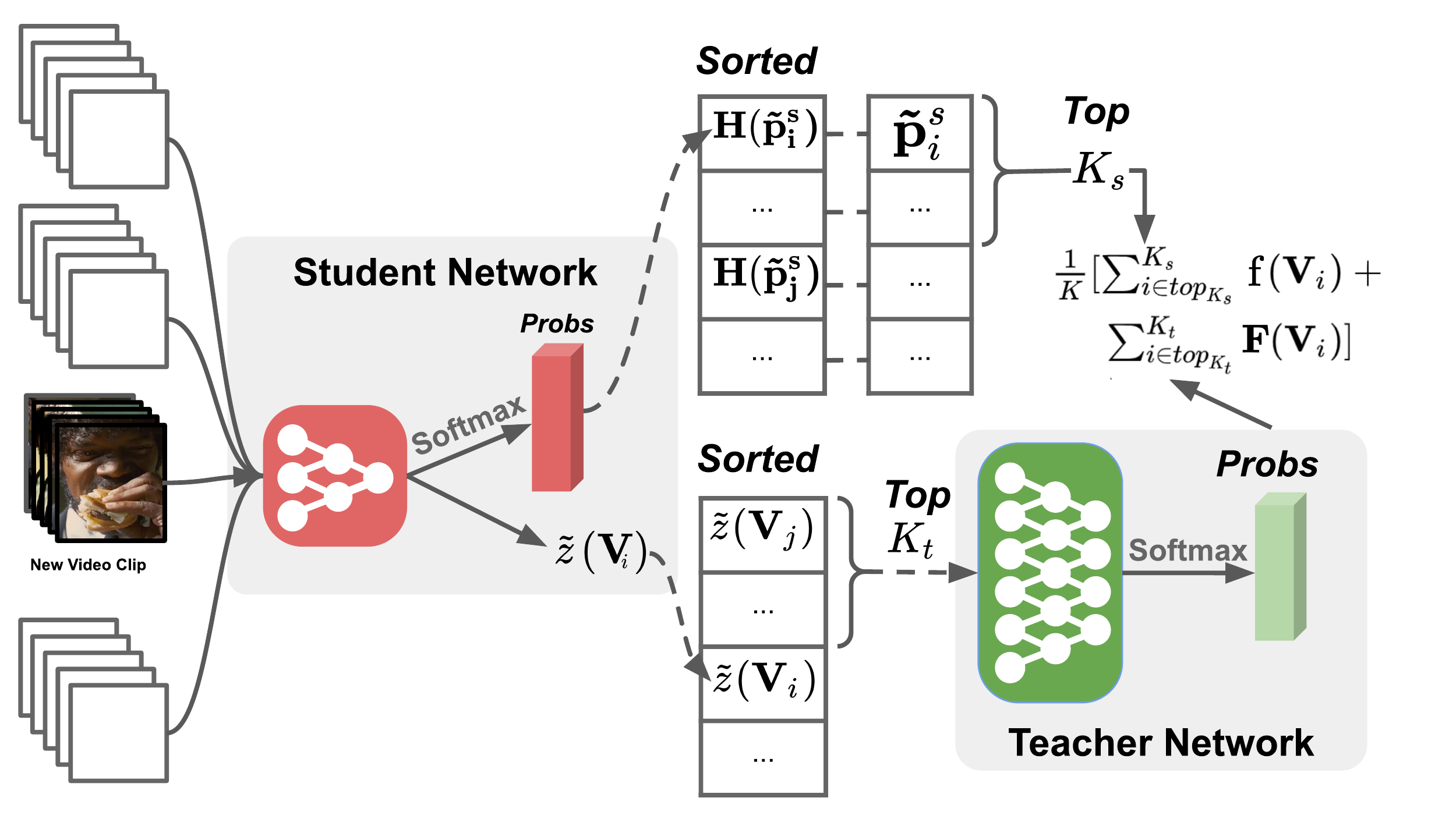}}}%
    \caption{In (a), the confidence distillation training schema from Sec. \ref{motivation} is visualized. For inference, in (b) the confidence scores $\tilde{z}$ are used to sort the top $K$ clips in descending order and the top $K$ are passed to $\mathrm{F}(\cdot)$ for classification. In (c), the predictive entropy of the student sampler's predictions is calculated over clips of $\mathcal{V}$. After sorting, only the top-$K_s$ prediction vectors are and averaged based on their predictive entropy from the student sampler; while the top-$K_t$ prediction vectors from the teacher are averaged based on confidence estimates.}%
    \label{fig:inference}
\end{figure*}

\section{Experiments}

{\bf Datasets:} The following datasets are considered to evaluate our approach: (1) the Kinetics dataset \cite{kay2017kinetics, carreira2018short, carreira2019short}, a video classification dataset generally used to determine a model's capacity to learn. (2) the UCF-101 dataset \cite{soomro2012ucf101}, one of the most widely used benchmark datasets; and (3) the Something-Something V2 dataset \cite{goyal2017something}, a dataset containing short trimmed video clips that show humans performing complex pre-defined actions with everyday objects. A characteristic of this dataset is that even though each video belongs to a single class, that class can be a combination of two other classes semantically. An overview of datasets can be seen in Table \ref{experiments:datasets}.

\begin{table}[h]
\begin{center}
{\resizebox{1\linewidth}{!}{
\begin{tabular}{l|c|z|c}
\toprule
\textsc{Dataset}                & \textsc{Instances} & \textsc{Avg. Length} & \textsc{Classes} \\ \hline
UCF-101                & 13320     & 7s              & 101     \\
Something-Something V2 & 220847    & 2-8s            & 174     \\
Kinetics               & Varies    & 10s             & 400-700 \\ 
\bottomrule
\end{tabular}
\vspace{-1mm}
}}
\end{center}
\caption{Overview of datasets used in our experiments.
\vspace{-0.1in}}
\label{experiments:datasets}
\vspace{-0.05in}
\end{table}

{\bf Models:} Two teacher backbones are considered: 3D residual networks (3D-ResNeXt with 101 layers from \cite{Hara_2018_CVPR} and spatio-temporal networks with 2+1D convolutions \cite{tran2018closer}. It is important to note that confidence distillation applies to most video-recognition models that use similar backbones. For the student architectures, we use 3D ShuffleNetV2 and MobileNetV2 \cite{kopuklu2019resource}. Model details are given in Table \ref{experiments:models}.

\begin{table}[h]
\begin{center}
{\resizebox{1\linewidth}{!}{
\begin{tabular}{ll|czc}
\toprule
&. & GFLOPs & Params & Depth \\ \hline
\multicolumn{1}{l|}{\textsc{Baseline}}               & 3D ResNeXt-101   & 6.932                       & 48.34M                      & 101                                 \\
\multicolumn{1}{l|}{}               & R(2+1)D          & 6.321                             & 38.76M                            & 34                                \\ \hline
\multicolumn{1}{l|}{\textsc{Sampler}}  & 3D-ShuffleNet-V2 & 0.360                       & 6.64M                      & 22                                 \\ 
\multicolumn{1}{l|}{}  & 3D-MobileNet-V2 & 0.446                       & 3.12M                      & 22                                 \\ 
\bottomrule
\end{tabular}
\vspace{-1mm}
}}
\end{center}
\caption{Models overview.\vspace{-0.05in}}
\label{experiments:models}
\vspace{-0.05in}
\end{table}

While both teacher models were designed for dense sampling, we show that the accuracy of these and similar state-of-the-art models can be improved compared to dense sampling by using confidence distillation.

{\bf Sampling baselines:} To sample $K$ clips from $\mathcal{V}$ we have the following trivial and non-trivial options: (1) Random sampling randomly choosing $K$ clips from a video (averaged over three runs). (2) Equidistant sampling chooses $K$ clips at equidistant time intervals. (3) Separate training (\textit{ST}) of a sampler without supervision signals by using predictive entropy (which we call \textit{ST-Ent}) or unsupervised confidence estimates \cite{devries2018learning} (which we call \textit{ST-Conf}). 

\textit{ST-Conf} utilizes an out-of-distribution detector with the intuition that ambiguous clips likely contain out-of-distribution features. $\mathrm{f}(\cdot)$ can be trained using the method in \cite{devries2018learning} to obtain unsupervised confidence estimates. The network's output has two branches, one for classification and one for confidence estimation. Given a clip $\mathbf{V}$ as an input, $\mathrm{f}_{st-conf}(\cdot)$ yields two vector for prediction and confidence logits. The confidence logit is then passed through a sigmoid function to obtain a confidence score $\tilde{z} \in [0,1]$. \textit{ST-Ent} utilizes predictive entropy \cite{shannon1948mathematical} that captures the average amount of information contained in the possible outcomes of a random variable. The higher the entropy, the more information the associated outcome has. This measure has been used \cite{gal2016uncertainty, hein2019relu} to capture uncertainty inherent in the data, such as an image of a number 7 that may appear similar to a number 1. To utilize \textit{ST-Ent}, we distill a student sampler using a soft-label distillation loss \cite{hinton2015distilling}. Other methods that train a model to do the video clip or frame sampling \cite{korbar2019scsampler, gao2020listen, wu2019adaframe} are complementary to ours and assume that the video is very long or redundant and the number of sampled clips is large; this prevents their usage in most video datasets. We do not assume the video is long or redundant. As such, \cite{gao2020listen, korbar2019scsampler, wu2019adaframe} can be stacked on top of our method as a divide and conquer strategy.

We construct an \textit{Oracle} sampler \cite{korbar2019scsampler} to calculate the upper bound of accuracy. \textit{Oracle} cheats by looking at the ground-truth label $y$ and only considers clips that yield the $k$ highest classification scores for $y$ to return an aggregate prediction.

\subsection{Implementation details}

{\bf Training:} We have implemented the models using PyTorch using four Nvidia P100 GPUs. Both the teacher and student take clip volumes of $16 \times 3 \times 112 \times 112$ as input. For $\lambda$ in Eq. \eqref{lossfunc}, we use $\lambda = \{0.5, 1.5, 2.0\}$ and $K \in \{1, 3, 5, 7, 10, 15\}$ where applicable.  We use $\tau \in \{ 0.9 \}$ for distillation. $\tau$ is typically \cite{hinton2015distilling} chosen to be a large number such as 5 or 20 in the beginning and may be annealed. We instead use much smaller $\tau$ values such that the teacher's entropy (information) will be dynamically adjusted depending on the student's confidence. We find that starting with smaller values as the baseline $\tau$ is better. We use $\tau = 0.9$ and see that higher values prevent the loss from converging.

We train the teacher and perform distillation for 50 epochs on the UCF-101 and Something-something V2 datasets and distill for 60 epochs on the Kinetics dataset. The confidence branch's learning rate starts the same as the rest of the network but decays step-wise, dividing by $5$ every epoch after the first epoch, while the learning rate of the rest of the network decays by dividing by $1.25$ every epoch. The models trained on Kinetics are trained similar to \cite{Hara_2018_CVPR} from scratch. The models trained on UCF-101 and Something-Something V2 are pre-trained on Kinetics and then fine-tuned for both the teacher and the student models. We find that for fine-tuning the classification branch of \textit{ConDi-SR} a learning rate $\alpha = 0.01$ works best and for training from scratch $\alpha = 0.1$ works better.

For Kinetics, The ResNeXt-101 teacher is trained on the Kinetics-600 dataset, and the R(2+1)D is trained on the Kinetics-400 dataset. UCF-101 and Something-Something V2 videos are too short for current non-trivial sampling baselines \cite{korbar2019scsampler, gao2020listen} to work effectively, and as such, they are not included for comparisons in these datasets.

To address the data imbalance of pseudo-ground-truth labels of Eq. \ref{psuedolabels} and prevent learning a trivial solution - i.e. uniform confidence scores for all inputs - we apply a higher weight $\mu$ is placed on the positive component of $\mathcal{L}_{\mathit{conf}}$. We find the positive weight multiplier $\mu = 1.5$ works well in all three datasets, while a larger value for $\mu$ prevented the distillation loss $\mathcal{L}_{KD}$ from decreasing. 
 
{\bf Evaluation:} We consider the Top-1 accuracy, meaning the actual class matches with the most probable class predicted by the model. For UCF-101, the mean values of test results are reported over the official validation folds. Since the test set annotations of Something-something V2 are unavailable, the validation set is used as the test set, and the training set is stratified $0.9/0.1$ into training and validation sets.

\subsection{Experimental results}

    


\begin{table*}[h]
\begin{center}
\resizebox{\textwidth}{!}{%
\begin{tabular}{lllllllllllll}
\toprule
\multicolumn{2}{l|}{} & \multicolumn{11}{c}{Dataset} \\ \cline{3-13} 
\multicolumn{2}{l|}{\multirow{-2}{*}{}} & \multicolumn{3}{c}{UCF-101} & \multicolumn{3}{c}{\cellcolor[HTML]{EFEFEF}SS v2} & \multicolumn{5}{c}{Kinetics (400/600)} \\ \hline
Baseline & \cellcolor[HTML]{EFEFEF}Teacher/Student & \multicolumn{1}{c}{K=1} & \multicolumn{1}{c}{K=5} & \multicolumn{1}{c}{Dense} & \multicolumn{1}{c}{\cellcolor[HTML]{EFEFEF}K=1} & \multicolumn{1}{c}{\cellcolor[HTML]{EFEFEF}K=3} & \multicolumn{1}{c}{\cellcolor[HTML]{EFEFEF}Dense} & \multicolumn{1}{c}{K=1} & \multicolumn{1}{c}{K=5} & \multicolumn{1}{c}{K=7} & \multicolumn{1}{c}{K=12} & \multicolumn{1}{c}{Dense} \\ \hline
Oracle & \cellcolor[HTML]{EFEFEF}ResNeXt & 95.9 & 92.81 & 89.08 & \cellcolor[HTML]{EFEFEF}50.1 & \cellcolor[HTML]{EFEFEF}43.02 & \cellcolor[HTML]{EFEFEF}35.1 & 84.8 & 78.9 & 76 & 70.37 & 68.3 \\
 & \cellcolor[HTML]{EFEFEF}R(2+1)D & 90 & 89.2 & 87 & \cellcolor[HTML]{EFEFEF}42.5 & \cellcolor[HTML]{EFEFEF}36.9 & \cellcolor[HTML]{EFEFEF}32.2 & 88.69 & 85 & 83.2 & 80.1 & 66.3 \\ \hline
Random (mean) & \cellcolor[HTML]{EFEFEF}ResNeXt & 86 & 88.98 & 89.08 & \cellcolor[HTML]{EFEFEF}24 & \cellcolor[HTML]{EFEFEF}30.24 & \cellcolor[HTML]{EFEFEF}35.1 & 52.1 & 65.7 & 65.9 & 66.02 & 68.3 \\
 & \cellcolor[HTML]{EFEFEF}R(2+1)D & 83.1 & 85.8 & 87 & \cellcolor[HTML]{EFEFEF}21.2 & \cellcolor[HTML]{EFEFEF}27.5 & \cellcolor[HTML]{EFEFEF}32.2 & 48.7 & 62.5 & 63.2 & 66.05 & 66.3 \\ \hline
Equidistant & \cellcolor[HTML]{EFEFEF}ResNeXt & 87.68 & 89.02 & 89.08 & \cellcolor[HTML]{EFEFEF}28.8 & \cellcolor[HTML]{EFEFEF}34.3 & \cellcolor[HTML]{EFEFEF}35.1 & 53.74 & 65.8 & 66.1 & 66.03 & 68.3 \\
 & \cellcolor[HTML]{EFEFEF}R(2+1)D & 84 & 85.9 & 87 & \cellcolor[HTML]{EFEFEF}27.75 & \cellcolor[HTML]{EFEFEF}30.6 & \cellcolor[HTML]{EFEFEF}32.2 & 50 & 62.6 & 63.2 & 66.02 & 66.3 \\ \hline
SC-Sampler \cite{korbar2019scsampler} & \cellcolor[HTML]{EFEFEF}R(2+1)D + ResNet-18 & \multicolumn{3}{c}{—} & \multicolumn{3}{c}{\cellcolor[HTML]{EFEFEF}—} & \multicolumn{3}{c}{—} & \textbf{70.9} & 66.3 \\ \hline
LTL Image-based \cite{gao2020listen} & \cellcolor[HTML]{EFEFEF}R(2+1)D + ResNet-18 & \multicolumn{3}{c}{73} & \multicolumn{3}{c}{\cellcolor[HTML]{EFEFEF}—} & \multicolumn{5}{c}{—} \\ \hline
BCE-Conf \cite{devries2018learning} & \cellcolor[HTML]{EFEFEF}ResNeXt + ShuffleNetV2 & 85.6 & 88.97 & 89.08 & \cellcolor[HTML]{EFEFEF}26 & \cellcolor[HTML]{EFEFEF}34.1 & \cellcolor[HTML]{EFEFEF}35.1 & 50 & 65.4 & 65.7 & 66 & 68.3 \\
 & \cellcolor[HTML]{EFEFEF}R(2+1)D + ShuffleNetV2 & 83 & 85.8 & 87 & \cellcolor[HTML]{EFEFEF}26.1 & \cellcolor[HTML]{EFEFEF}33.5 & \cellcolor[HTML]{EFEFEF}32.2 & 48.7 & 62.5 & 63.2 & 66.05 & 66.3 \\
 & \cellcolor[HTML]{EFEFEF}ResNeXt + MobileNetV2 & 85.5 & 88.95 & 89.08 & \cellcolor[HTML]{EFEFEF}26 & \cellcolor[HTML]{EFEFEF}34.05 & \cellcolor[HTML]{EFEFEF}35.1 & 48.45 & 63.11 & 64.13 & 67 & 68.3 \\
 & \cellcolor[HTML]{EFEFEF}R(2+1)D + MobileNetV2 & 82.9 & 85.9 & 87 & \cellcolor[HTML]{EFEFEF}26.2 & \cellcolor[HTML]{EFEFEF}33.51 & \cellcolor[HTML]{EFEFEF}32.2 & 47.6 & 61.25 & 61.9 & 66 & 66.3 \\ \hline
ST-Ent \cite{gal2016uncertainty} & \cellcolor[HTML]{EFEFEF}ResNeXt + ShuffleNetV2 & 87.6 & 89.1 & 89.08 & \cellcolor[HTML]{EFEFEF}32.6 & \cellcolor[HTML]{EFEFEF}34.9 & \cellcolor[HTML]{EFEFEF}35.1 & 60.8 & 65.8 & 66.3 & 65.95 & 68.3 \\
 & \cellcolor[HTML]{EFEFEF}R(2+1)D + ShuffleNetV2 & 85.5 & 86.9 & 87 & \cellcolor[HTML]{EFEFEF}28.8 & \cellcolor[HTML]{EFEFEF}31.2 & \cellcolor[HTML]{EFEFEF}32.2 & 58.72 & 65 & 66.2 & \textit{66.4} & 66.3 \\
 & \cellcolor[HTML]{EFEFEF}ResNeXt + MobileNetV2 & 87.54 & 89.09 & 89.08 & \cellcolor[HTML]{EFEFEF}33 & \cellcolor[HTML]{EFEFEF}35 & \cellcolor[HTML]{EFEFEF}35.1 & 57.11 & 64.71 & 66.1 & 66.9 & 68.3 \\
 & \cellcolor[HTML]{EFEFEF}R(2+1)D + MobileNetV2 & 85.1 & 86.7 & 87 & \cellcolor[HTML]{EFEFEF}28.76 & \cellcolor[HTML]{EFEFEF}31.3 & \cellcolor[HTML]{EFEFEF}32.2 & 56.4 & 64.64 & 66.07 & 66.1 & 66.3 \\ \hline
ST-Conf & \cellcolor[HTML]{EFEFEF}ResNeXt + ShuffleNetV2 & 86.8 & 88.9 & 89.08 & \cellcolor[HTML]{EFEFEF}30 & \cellcolor[HTML]{EFEFEF}34.4 & \cellcolor[HTML]{EFEFEF}35.1 & 55.05 & 65.35 & 65.7 & 66.02 & 68.3 \\
 & \cellcolor[HTML]{EFEFEF}R(2+1)D + ShuffleNetV2 & 84 & 86.1 & 87 & \cellcolor[HTML]{EFEFEF}26.9 & \cellcolor[HTML]{EFEFEF}30.6 & \cellcolor[HTML]{EFEFEF}32.2 & 58.63 & 65.6 & 66 & 66.2 & 66.3 \\
 & \cellcolor[HTML]{EFEFEF}ResNeXt + MobileNetV2 & 86.7 & 89 & 89.08 & \cellcolor[HTML]{EFEFEF}30.3 & \cellcolor[HTML]{EFEFEF}34.3 & \cellcolor[HTML]{EFEFEF}35.1 & 54.8 & 65.12 & 65.7 & 66.01 & 68.3 \\
 & \cellcolor[HTML]{EFEFEF}R(2+1)D + MobileNetV2 & 88.8 & 87.1 & 87 & \cellcolor[HTML]{EFEFEF}26.92 & \cellcolor[HTML]{EFEFEF}30.61 & \cellcolor[HTML]{EFEFEF}32.2 & 57.7 & 65.1 & 66 & 66.2 & 66.3 \\ \hline
\textbf{ConDi-SR (ours)} & \cellcolor[HTML]{EFEFEF}ResNeXt + ShuffleNetV2 & \textbf{89} & \textbf{91.2} & 89.08 & \cellcolor[HTML]{EFEFEF}35.7 & \cellcolor[HTML]{EFEFEF}38.33 & \cellcolor[HTML]{EFEFEF}35.1 & \textbf{65.1} & \textbf{70.8} & \textbf{71.25} & 69.8 & 68.3 \\
 & \cellcolor[HTML]{EFEFEF}R(2+1)D + ShuffleNetV2 & \textbf{86.06} & \textbf{87.4} & 87 & \cellcolor[HTML]{EFEFEF}31.13 & \cellcolor[HTML]{EFEFEF}34 & \cellcolor[HTML]{EFEFEF}32.2 & \textbf{65.08} & \textbf{71.15} & \textbf{71.2} & 69.9 & 66.3 \\
 & \cellcolor[HTML]{EFEFEF}ResNeXt + MobileNetV2 & 88.8 & 91 & 89.08 & \cellcolor[HTML]{EFEFEF}\textbf{36.04} & \cellcolor[HTML]{EFEFEF}\textbf{38.35} & \cellcolor[HTML]{EFEFEF}35.1 & 62.3 & 68.7 & 69.3 & 68.4 & 68.3 \\
 & \cellcolor[HTML]{EFEFEF}R(2+1)D + MobileNetV2 & 86.03 & 87.3 & 87 & \cellcolor[HTML]{EFEFEF}\textbf{31.66} & \cellcolor[HTML]{EFEFEF}\textbf{34.2} & \cellcolor[HTML]{EFEFEF}32.2 & 61.1 & 68.9 & 69.3 & 68.3 & 66.3 \\
\bottomrule
\end{tabular}%
}
\caption{The best results are shown in bold for each column separately. As $K$ increases, all methods converge to the same point representing dense sampling. While equidistant sampling shows to be highly effective in smaller videos, as the number of clips in the video increase, its accuracy drops significantly (c, f), while our method remains consistently 10-15\% higher in terms of accuracy in longer videos. It is also worth noting that for a smaller $K$, our method outperforms all other baselines.}
\label{tab:classification_accuracy}
\end{center}
\vspace{-15pt}
\end{table*}

\cite{devries2018learning}
\cite{gal2016uncertainty}
\cite{gao2020listen}
\cite{korbar2019scsampler}

{\bf Accuracy and efficiency:} To be comparable to baselines, in this section we use type (a) inferencing from Fig. \ref{fig:inference}; experimental results with classification accuracy are shown in Fig. \ref{tab:classification_accuracy}. The Kinetics dataset is big enough to train models from scratch. However, there are different variants, and a percentage of video samples may not be available or accessible. Nevertheless, we distill a student using this dataset and report the results. Due to the lack of available teacher models, the ResNeXt-101 is trained on the Kinetics-600 dataset, and the R(2+1)D is trained on the Kinetics-400 dataset. We compare the non-trivial baselines \cite{korbar2019scsampler, devries2018learning} and it is important to note that the results reported directly from \cite{korbar2019scsampler} have access to the audio while our method does not require audio access. The videos from UCF-101 and Something-something v2 are too short for other state-of-the-art redundancy samplers \cite{korbar2019scsampler, gao2020listen} to work effectively, leaving \cite{devries2018learning} as the only applicable baseline. 

\begin{table}[h]
\begin{center}
{\resizebox{1\linewidth}{!}{
\begin{tabular}{ll|cccc}
\toprule
&       & \multicolumn{1}{c|}{\textbf{ConDi-SR (ours)}}               & \multicolumn{1}{c|}{Equidistant}   & \multicolumn{1}{l|}{BCE-Conf \cite{devries2018learning}}      & \multicolumn{1}{l}{BCE-Ent \cite{gal2016uncertainty}} \\ \hline
\multicolumn{1}{l|}{Dataset}       & K     & \multicolumn{4}{c}{Mean time per video / (Accuracy)}                                                                                                \\ \hline
\multicolumn{1}{l|}{\textsc{UCF}}      & 1     & \multicolumn{1}{c|}{1.24s (88.87)}          & \multicolumn{1}{c|}{0.29s (86.5)}  & \multicolumn{1}{c|}{1.24s (85.67)} & 1.2s (87.06)                \\
\multicolumn{1}{l|}{101}             & 3     & \multicolumn{1}{c|}{1.9s (90.12)}           & \multicolumn{1}{c|}{0.84s (88.9)}  & \multicolumn{1}{c|}{1.9s (87.92)}  & 1.84s (89.01)               \\
\multicolumn{1}{l|}{}             & 5     & \multicolumn{1}{c|}{\textbf{2.31s (91.28)}} & \multicolumn{1}{c|}{1.42s (89.5)}  & \multicolumn{1}{c|}{2.31s (89.43)} & 2.01s (89.7)                \\
\multicolumn{1}{l|}{}             & 7     & \multicolumn{1}{c|}{3.1s (90.45)}           & \multicolumn{1}{c|}{2.1s (89.6)}   & \multicolumn{1}{c|}{3.1s (89.74)}  & 2.9s (89.8)                 \\ \cline{2-6} 
\multicolumn{1}{l|}{}             & All & \multicolumn{4}{c}{3.7s (89.8)}                                                                                                                     \\ \hline
\multicolumn{1}{l|}{\textsc{Kinetics}} & 1     & \multicolumn{1}{c|}{1.31s (65.18)}          & \multicolumn{1}{l|}{0.29s (54.51)} & \multicolumn{1}{c|}{1.31s (50.1)}  & 1.2s (60.9)                 \\
\multicolumn{1}{l|}{600}             & 3     & \multicolumn{1}{c|}{1.89s (67.30)}          & \multicolumn{1}{l|}{0.84s (60.12)} & \multicolumn{1}{c|}{1.89s (63.14)} & 1.84s (64.3)                \\
\multicolumn{1}{l|}{}             & 5     & \multicolumn{1}{c|}{2.28s (69.37)}          & \multicolumn{1}{l|}{1.42s (66.2)}  & \multicolumn{1}{c|}{2.28s (64.9)}  & 2.02s (65.62)               \\
\multicolumn{1}{l|}{}             & 7     & \multicolumn{1}{c|}{\textbf{3.05s (71.47)}} & \multicolumn{1}{l|}{2.1s (67.11)}  & \multicolumn{1}{c|}{3.05s (66.16)} & 2.91s (66.3)                \\ \cline{2-6} 
\multicolumn{1}{l|}{}             & All & \multicolumn{4}{c}{5.3s (68.3)} \\                        \bottomrule
\end{tabular}
}}
\caption{Computational Savings @ different $K$ values for ResNeXt-101 teacher model and ShuffleNetV2 sampler.}
\label{tab:speed}
\end{center}
\vspace{-15pt}
\end{table}

{\bf Division of workload:} $\mathrm{f}(\cdot)$ uses at least an order of magnitude fewer floating-point operations than that of $\mathrm{F}(\cdot)$. As shown in Fig. \ref{fig:inference} (b), we can divide the workload using \textit{ConDi-SR} by selecting $K_t$ and $K_s$ clips for the teacher and student to predict respectively where $K_t + K_s = K$. Results on UCF-101 and Kinetics-600 utilizing a ResNeXt-101 teacher are shown in Table \ref{tab:divided}.

\begin{table}[h]
\begin{center}
{\resizebox{1\linewidth}{!}{
\begin{tabular}{l|czc}
\toprule
Dataset            & $K_s$ & Accuracy & Mean time per video (s) \\ \hline
\textsc{UCF-101} (K = 5)    & 0     & 91.2     & 2.31                    \\
                   & 1     & 90.65    & 2.16                    \\
                   & 3     & 90.105   & 1.96                    \\
                   & 5     & 89.03    & 1.7                     \\ \hline
\textsc{Kinetics-600} (K=7) & 0     & 71.5     & 3.05                    \\
                   & 1     & 70.99    & 2.9                     \\
                   & 4     & 69.51    & 2.6                     \\
                   & 7     & 66.89    & 2.17                    \\ 
                   \bottomrule

\end{tabular}
}}
\vspace{2mm}
\caption{Dividing the workload between the teacher and student at different $K_s$ values for the ResNeXt-101 teacher model and the ShuffleNetV2 student sampler; This approach will be more useful when a video has more clips (i.e. $K$ is large).}
\label{tab:divided}
\end{center}
\vspace{-5mm}
\end{table}

\section{Conclusion}

We proposed the confidence distillation framework and empirically showed its ability to increase efficiency and accuracy in classifying actions across three diverse datasets and utilized the learned shared representation to divide the workload between the student and the teacher optionally. We also verified that using our method, accuracy and efficiency can be improved across a variety of student and teacher backbone combinations. Furthermore, confidence distillation can be utilized for both compact and lengthy videos, unlike all other state-of-the-art sampling methods. For future work, training the student sampler on other modalities such as audio or optical flow or extending our method to other machine learning tasks such as out of distribution detection are noted.

\newpage

\bibliographystyle{IEEEtran}
\bibliography{IEEEabrv, IEEEexample}
\end{document}